\documentclass[journal]{IEEEtran}

\usepackage{amsmath,amsfonts}
\usepackage{algorithmic}
\usepackage{algorithm}
\usepackage{array}
\usepackage[caption=false,font=normalsize,labelfont=sf,textfont=sf]{subfig}
\usepackage{textcomp}
\usepackage{stfloats}
\usepackage{url}
\usepackage{verbatim}
\usepackage{graphicx}
\usepackage{placeins}
\usepackage{cleveref}
\usepackage{cite}
\usepackage{adjustbox} 
\usepackage{threeparttable}
\usepackage{booktabs} 

\begin{document}

\title{A Data-Centric Perspective on the Influence of Image Data Quality in Machine Learning Models}

\author{
    \IEEEauthorblockN{Pei-Han Chen and Szu-Chi Chung$^*$} \\
    \IEEEauthorblockA{
        Department of Applied Mathematics, National Sun Yat-sen University, Kaohsiung, Taiwan
    }
    \thanks{* Corresponding author: steve2003121@gmail.com}
    \thanks{The authors acknowledge the support of the National Science and Technology Council, Taiwan, under Grant 114-2813-C-110-042-E.}
}



\maketitle

\begin{abstract}
In machine learning, research has traditionally focused on model development, with relatively less attention paid to training data. As model architectures have matured and marginal gains from further refinements diminish, data quality has emerged as a critical factor. However, systematic studies on evaluating and ensuring dataset quality in the image domain remain limited.

This study investigates methods for systematically assessing image dataset quality and examines how various image quality factors influence model performance. Using the publicly available and relatively clean CIFAKE dataset, we identify common quality issues and quantify their impact on training. Building on these findings, we develop a pipeline that integrates two community-developed tools, CleanVision and Fastdup. We analyze their underlying mechanisms and introduce several enhancements, including automatic threshold selection to detect problematic images without manual tuning.

Experimental results demonstrate that not all quality issues exert the same level of impact. While convolutional neural networks show resilience to certain distortions, they are particularly vulnerable to degradations that obscure critical visual features, such as blurring and severe downscaling. To assess the performance of existing tools and the effectiveness of our proposed enhancements, we formulate the detection of low-quality images as a binary classification task and use the F1 score as the evaluation metric. Our automatic thresholding method improves the F1 score from 0.6794 to 0.9468 under single perturbations and from 0.7447 to 0.8557 under dual perturbations. For near-duplicate detection, our deduplication strategy increases the F1 score from 0.4576 to 0.7928. These results underscore the effectiveness of our workflow and provide a foundation for advancing data quality assessment in image-based machine learning.
\end{abstract}


\section{Introduction}

Data-centric artificial intelligence (DCAI) is an emerging paradigm that focuses on improving model performance by enhancing the quality of training data, rather than relying solely on innovations in model architecture. In the field of image processing, high-quality, diverse, and accurately labeled image data are often the key determinants of system performance. Traditional model-centric approaches typically emphasize tuning model parameters and architectures but tend to overlook potential issues within the data itself — such as annotation errors, dataset bias, and noise. In contrast, DCAI advocates for proactively improving data quality through systematic techniques, including data cleaning, augmentation, annotation correction, and data selection — thus providing a more robust foundation for model training~\cite{renggli2021data}.

In recent years, as machine learning models have become increasingly mature, the performance gains from architectural improvements have gradually diminished. Consequently, the research community has begun to shift its focus toward the quality of training data. However, the literature still lacks substantial research on methods for ensuring the quality of image datasets. In fact, even widely used community datasets such as MNIST and CIFAR-10 have been shown to contain labeling errors~\cite{northcutt2021pervasive}, which may adversely affect both the training process and the model's generalization capabilities. Meanwhile, many existing image-cleaning workflows continue to rely on subjective human judgment to identify and remove problematic data — or, in some cases, forgo additional processing altogether. While such practices may still yield reasonably good results, model performance remains limited, even with extensive parameter tuning or increased training iterations.

To address these challenges, this study investigates how various aspects of image quality influence model performance. Through a series of quantitative experiments, we seek to identify the most influential factors. Based on these findings, we examine algorithmic principles proposed in prior literature and evaluate open-source tools~\cite{mazumder2023dataperf,MLCommonsWebsite,DynabenchWebsite,kuan2022back,de2009character} for their ability to detect image quality issues. Building upon the insights and methodologies from these tools, we propose a systematic pipeline for image dataset cleaning and demonstrate its improved performance over existing approaches. We believe the results of this study are promising and pave the way toward automatically and accurately flagging problematic images in large-scale datasets.

\section{Related works}
Artificial Intelligence (AI) is undergoing a paradigm shift — from a model-centric to a data-centric approach. While the past decade emphasized innovations in model architectures and algorithms \cite{jakubik2024data, zha2025data}, attention has increasingly turned to data quality and engineering as key drivers of performance and reliability. This transition has given rise to DCAI, defined as the systematic engineering of data used in AI systems. DCAI posits that, for many tasks, the model is no longer the bottleneck; rather, iterative improvements in data quality yield the most significant performance gains \cite{jakubik2024data}. In this view, data is treated as a dynamic asset — akin to code — that requires versioning, management, and continuous refinement. As a result, the focus has shifted from simply amassing large datasets to curating high-quality ones. Deep learning performance is fundamentally dependent on data quality \cite{smith2024quality, steinhauser2025data}, and low-quality images — affected by issues such as poor brightness, contrast, color distortion, noise, and blur — directly hinder a model's ability to extract meaningful features.

A core pillar of DCAI is dataset curation: assembling clean, balanced, and task-representative datasets. The DataPerf benchmark suite was proposed to standardize dataset evaluation and promote improvements in data quality \cite{mazumder2023dataperf}. Studies have also shown that improving data quality can surpass the performance gains obtained from using more complex models. For example, Northcutt et al. demonstrated that correcting mislabeled test samples enabled smaller models to outperform deeper ones, revealing the hidden cost of label noise \cite{northcutt2021pervasive}. Another key area of focus in DCAI is image enhancement and denoising. In real-world scenarios, noise from low-light conditions, sensor artifacts, or motion blur can severely degrade visual features. Competitions such as NTIRE have introduced real noisy/clean image pairs to benchmark denoising algorithms aimed at improving both perceptual quality and downstream task performance \cite{Abdelhamed2020NTIRE}. Deep learning has revitalized image denoising, a long-standing task in image processing \cite{elad2023image}. For instance, DnCNN learns a residual mapping from noisy inputs to the noise component \cite{zhang2017beyond}, while FFDNet incorporates a noise-level map to efficiently handle varying noise conditions \cite{zhang2018ffdnet}. Image deblurring — recovering sharp images from blurred inputs — has also greatly benefited from deep learning \cite{zhang2022deep}.

In addition to improving image quality, researchers have explored methods for detecting low-quality or redundant samples within datasets. Duplicates, irrelevant content, and artifacts (e.g., watermarks or diagrams) reduce data diversity and hinder learning. Tools such as CleanVision detect various anomalies, including blur, darkness, and near-duplicates \cite{cleanvision}. When applied to datasets like Caltech-256 and CIFAR-10, CleanVision has uncovered widespread issues such as mislabeled diagrams. Similarly, deduplication studies on CLIP have shown that removing duplicates can improve model performance \cite{mayilvahanan2023does}. Another strategy involves feature-space outlier detection, which flags samples that deviate from the main distribution — either removing noise or isolating rare but valuable edge cases \cite{kuan2022back}. These approaches align with the broader concept of dataset debugging, where eliminating low-value samples enhances learning clarity and generalization.

Despite growing interest in DCAI, few studies provide quantitative assessments of how specific types of image degradation affect model training. Moreover, automatically flagging low-quality images remains a challenging task. This study aims to fill that gap by systematically evaluating the impact of common image degradations on Convolutional Neural Network (CNN) performance and identifying critical factors that significantly influence learning. While tools such as CleanVision and Fastdup \cite{Fastdup} are widely used to detect problematic samples — including duplicates, blurs, brightness anomalies, and aspect ratio inconsistencies — their internal mechanisms remain underexplored. This study also examines how these tools compute image similarity, detect anomalies, and implement clustering or outlier detection. Our goal is to advance not only dataset quality but also our understanding of the algorithmic principles underlying modern dataset auditing tools.

Finally, these tools often require manual tuning of hyperparameters — particularly the threshold for flagging problematic images — a process that is both subjective and labor-intensive. To address this limitation, we propose an automated pipeline that integrates and improves upon these tools by refining their underlying algorithms. We further introduce a threshold selection strategy designed to automate image quality assessment, enabling a fully user-free dataset cleaning pipeline. By developing principled methodologies for identifying and mitigating quality issues in image datasets, this study supports the broader adoption of DCAI in real-world machine learning workflows.

\section{The Proposed Methodologies}
In this section, we provide a detailed overview of the algorithmic principles underlying CleanVision and Fastdup, as well as our proposed pipeline and improvements.

\subsection{CleanVision Testing and Algorithm}
This subsection summarizes our investigation of the CleanVision package. CleanVision is designed to detect nine types of anomalous data: light, dark, blurry, low-information, odd size (significant deviations in image dimensions), odd aspect ratio (extreme variations in aspect ratio), grayscale, and exact or near duplicates.

The workflow of the program can be broadly divided into five steps:
\begin{enumerate}
    \item \textbf{Load images.}
    \item \textbf{Specify issue types:} Determine which issue types to detect (all are enabled by default).
    \item \textbf{Compute scores:} Invoke the detection function to calculate a score for each specified issue, with values ranging from $[0,1]$, where lower scores indicate a higher likelihood of being problematic.
    \item \textbf{Thresholding:} Compare scores against a fixed threshold (either default or user-defined) to identify anomalous images. Some issue types do not require thresholds; for example, grayscale and duplicate detections are flagged whenever their scores differ from 1.
    \item \textbf{Output results:} Store the detection results in a dataframe, summarize the number of low-quality images, and visualize representative examples for user inspection.
\end{enumerate}

For light and dark detection, the scoring algorithm proceeds as follows:
\begin{enumerate}
    \item If the image is in RGB format, it is first converted to grayscale using a human-vision–based formula,
    \[
        \sqrt{0.241R^2 + 0.691G^2 + 0.068B^2},
    \]
    which is similar to the method used in Photoshop\footnote{\url{https://alienryderflex.com/hsp.html}}.
    \item Each pixel value is normalized by dividing by 255, and the 1st, 5th, 10th, 15th, 90th, 95th, and 99th percentiles, along with the mean brightness, are computed.
    \item The dark score is defined as the 99th percentile value, whereas the light score is defined as $1$ minus the 5th percentile value.
\end{enumerate}

Blurry image detection is conducted by first converting images to grayscale. The algorithm combines two complementary metrics: edge detection using Laplacian variance (which measures the sharpness of edges in the image), and grayscale histogram standard deviation (which captures texture variation). The Laplacian variance is computed as the variance of the resulting edge image. Images with lower combined scores indicate more severe blurriness.

Odd-size images are detected using a statistical outlier detection approach based on the Interquartile Range (IQR) method. The algorithm first calculates each image's size as the square root of its area ($\text{width} \times \text{height}$), then computes the 25th and 75th percentiles (Q1, Q3) of all image sizes in the dataset to establish minimum and maximum thresholds using
\begin{align*}
    \text{min\_threshold} &= Q1 - \text{iqr\_factor} \times \text{IQR}, \\
    \text{max\_threshold} &= Q3 + \text{iqr\_factor} \times \text{IQR}.
\end{align*}
with a default IQR factor of 3.0. For each image, it calculates the distance from the midpoint between thresholds, normalizes this distance, and converts it to a score between 0 and 1. Images scoring below the threshold are flagged as having odd size issues.

For detecting odd aspect ratios, the algorithm computes an aspect ratio score using $ \min(\text{width}/\text{height},\ \text{height}/\text{width})$ to capture extreme shapes regardless of orientation. Images are flagged as having odd aspect ratios when their score falls below a certain threshold.

The grayscale score is binary (0 or 1). The package first attempts to determine the image mode using PIL~\cite{pillow}. If this fails, the image is converted into a NumPy array~\cite{harris2020array}. When the array is two-dimensional, or when the three channel values are identical, the mode is set to “L” (8-bit grayscale). If the mode is “L”, the grayscale score is set to 0.

For duplicate detection, a hash-based method is employed: MD5 hashing is used for exact duplicates, and perceptual hashing (pHash) is used for near duplicates. Each image is assigned a hash value, and images sharing the same hash are grouped together. The score for each image is then defined as:
\[
    \frac{1}{\text{len(set to which the image belongs)}}.
\]
Images with scores smaller than 1 are flagged as duplicates.

\subsection{Fastdup Testing and Algorithm}
Fastdup can detect several types of anomalies, including invalid images, duplicate image pairs, outliers, dark, bright, and blurry images, as well as image clusters. In this study, we focus primarily on the detection categories that overlap with CleanVision.

After loading the images, Fastdup identifies duplicate image pairs by extracting feature vectors using a proprietary pre-trained Open Neural Network Exchange (ONNX) model. A nearest-neighbor search is then performed to find the two most similar images for each sample, and cosine distances are computed and normalized to the $[0,1]$ range, where a value of 1 indicates identical images. Pairs with cosine similarity greater than 0.96 are considered near-duplicates and are grouped using a connected-component algorithm.

For dark and bright image detection, Fastdup computes the mean pixel value of each image. From the statistical files generated by Fastdup, we obtained the per-channel RGB means. We then applied a linear regression model in \texttt{scikit-learn}~\cite{pedregosa2011scikit} and computed the overall brightness as $(R + G + B)/3$, comparing it with the reported mean. Based on this comparison, we infer that Fastdup's overall mean is calculated as the average of the per-channel RGB means. Dark images are ranked from lowest to highest mean value, while bright images are ranked from highest to lowest.

Blurry images are detected by applying the Laplacian operator to compute the variance, similar to the approach used in CleanVision. Images with lower variance are more likely to be blurry and are ranked in ascending order of variance for user inspection. Although thresholds for anomaly detection are defined within the package, they are not directly applied in the current implementation.

\begin{figure*}
    \centering
    \includegraphics[width=\linewidth]{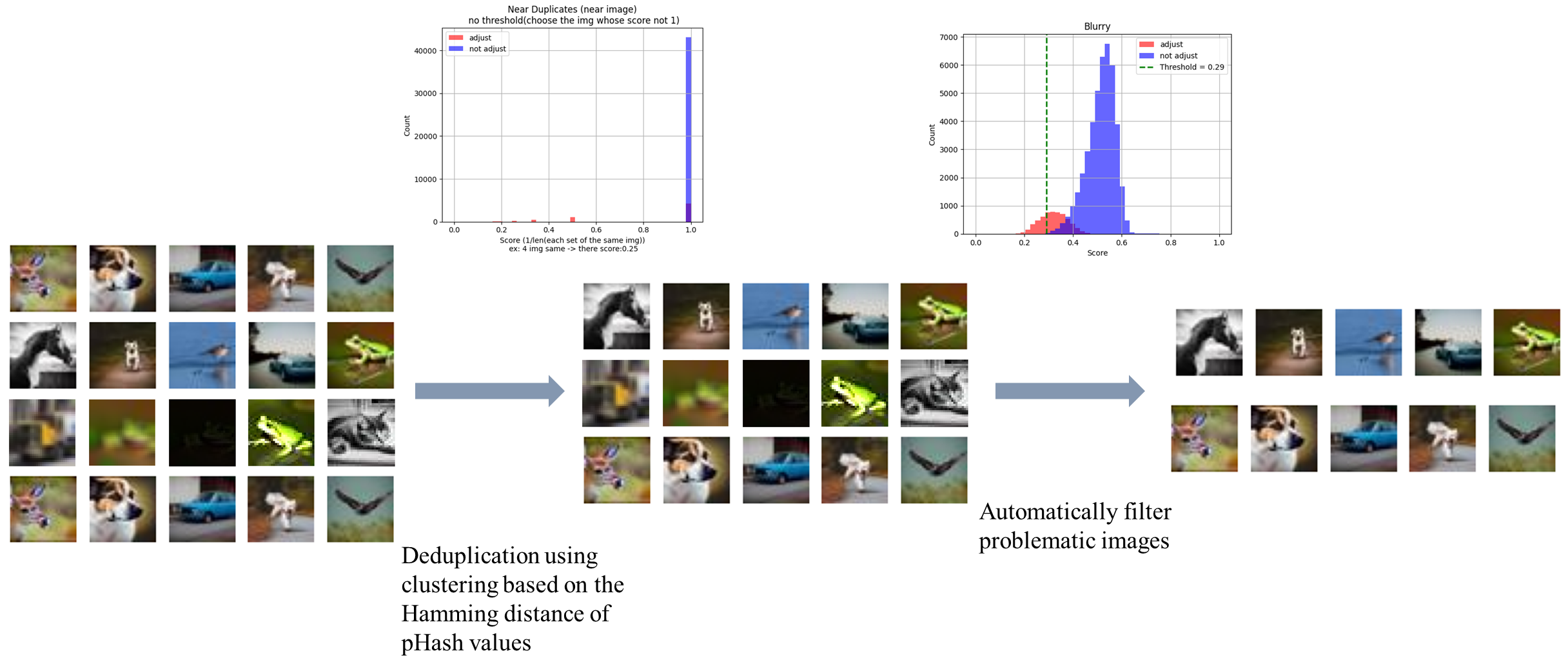}
    \caption{The proposed workflow performs deduplication by applying clustering based on the Hamming distance of pHash values. In addition, it computes quality scores using either CleanVision or Fastdup and applies automatic threshold selection to identify low-quality images for user inspection and filtering.}
    \label{fig:workflow}
\end{figure*}

\subsection{Proposed Workflow}
Our proposed workflow builds upon the capabilities of CleanVision and Fastdup, integrating the strengths of both packages. In addition, we modified the detection mechanisms for light, grayscale, and near-duplicate images as described below.

For light detection, under the default configuration, the overlap between the score distributions of adjusted and unadjusted images was substantial in our experiments, leading to poor discrimination. The original grayscale conversion formula used in the algorithm was nonstandard, so we tested several alternatives, including $0.2126R + 0.7152G + 0.0722B$, $0.299R + 0.587G + 0.114B$, $\sqrt{0.299R^2 + 0.587G^2 + 0.114B^2}$, and $\frac{R + G + B}{3}$. The results were largely comparable, with the original formula performing slightly better; thus, no change was made. However, the original scoring method, which relies on $1 - \text{5th percentile}$, tended to assign excessively high scores to overexposed images that still contained sufficiently dark regions, making the 5th-percentile criterion overly stringent. To address this, we computed additional percentiles (25th, 30th, 40th, 50th, 60th, and 75th). Among these, the 60th and 75th percentiles yielded better discrimination performance.

For grayscale detection, in the original implementation, the check for whether an image is a two-dimensional array or has identical values across all three channels occurs after mode determination. Consequently, RGB images with identical channel values were not detected as grayscale. To fix this issue, we modified the process so that the check for identical channel values is performed after PIL determines the mode, even when the mode is not "L." With this enhanced re-detection, we discovered several grayscale images in the original dataset that had previously gone unnoticed.

For near-duplicate detection, in CleanVision, the number of near-duplicate images detected using the default configuration was limited. We inferred that the strict requirement for identical hash values to group images was overly restrictive. Because the perceptual hash (pHash) values of two images become increasingly similar as the images themselves become more alike, we incorporated Fastdup's distance-based similarity concept and experimented with clustering methods. Specifically, we applied HDBSCAN and hierarchical clustering using the Hamming distance of the pHash values, while tuning parameters such as the linkage criterion in the hierarchical clustering strategy. We also evaluated the effect of dimensionality reduction by testing different methods, both with and without applying them before clustering. We found that hierarchical clustering with the \texttt{linkage} parameter set to \texttt{single} achieved the best results \footnote{For near-duplicate detection in Fastdup, we first used the default feature vectors produced by its pre-trained model. We then tested clustering with HDBSCAN using cosine and Euclidean distances, adjusting the \texttt{min\_cluster\_size} parameter, and applying or omitting PCA for dimensionality reduction and normalization of feature vectors. We also experimented with hierarchical clustering and alternative built-in models that generate lower-dimensional feature vectors. However, none of these approaches yielded satisfactory results.}.

Finally, in our proposed workflow, we suggest to combine both approaches for duplicate detection. CleanVision focuses on detecting similarity at the pixel level, while Fastdup operates in the latent space to identify semantic similarity. By integrating these two complementary methods, our workflow can capture both pixel-level and semantic-level redundancies, thereby improving the robustness of duplicate image detection.

\subsection{Thresholding Algorithms}
The original thresholds in both CleanVision and Fastdup are hard-coded constants, making them unsuitable for application across different datasets. To address this limitation, we employed the improved detection methods to evaluate the performance of several thresholding algorithms.

In our experiments, we observed that the score distributions of low-quality and normal images generally exhibited a bimodal pattern. This observation motivated the exploration of several histogram-based thresholding algorithms, many of which are commonly used in image binarization tasks. These include Otsu's Method (Otsu), which maximizes inter-class variance~\cite{otsu1975threshold}; Kittler–Illingworth Minimum Error Thresholding (MET), which minimizes the average pixel classification error rate~\cite{kittler1986minimum}; Li's Minimum Cross-Entropy Thresholding (Li), which minimizes the cross-entropy between two classes~\cite{li1993minimum}; and Maximum Entropy Thresholding (Max Entropy), which maximizes the sum of the entropies of the two classes~\cite{kapur1985new}.

In addition, we considered Generalized Histogram Thresholding (GHT), which unifies Otsu, MET, and Weighted Percentile Thresholding~\cite{barron2020generalization}; the Modified Valley Emphasis method (MVE), which enhances Otsu's approach~\cite{xing2020automatic}; and the Gamma Mixture Model (GMM), which fits two gamma distributions using a maximum likelihood estimation procedure~\cite{nouri2018optimized}. Although GMM is commonly associated with Gaussian distributions, prior work~\cite{nouri2018optimized} reported that modeling with gamma distributions produces better results. Therefore, the algorithm is referred to as GMM throughout this study.

\section{Experiment Results}

\subsection{The Impact of Data Quality on CNN Model Accuracy}
\label{sec:impac}

Our experimental investigation began with the FAKE subset of the CIFAKE dataset~\cite{bird2024cifake}, which was generated using Stable Diffusion \cite{rombach2022high}. It comprises 50,000 training and 10,000 testing images spanning 10 different classes, each with a resolution of $32 \times 32$ pixels. As the dataset consists of AI-generated images, it is assumed to be of high initial quality. From the training data, 5,000 images were set aside as a validation set, leaving 45,000 images for training.

The baseline model — a compact CNN whose architecture is detailed in \Cref{tbl:model_arch} — was trained on this clean dataset. This configuration achieved a classification accuracy of 88.52\% on the unseen test set, establishing a benchmark for subsequent experiments.

To simulate real-world data quality issues, we systematically introduced a series of perturbations to the entire training dataset using OpenCV~\cite{bradski2000opencv} and Albumentations~\cite{buslaev2020albumentations}. The CNN was retrained from scratch on each degraded dataset to isolate the effect of each factor on classification accuracy. The specific perturbations were defined as follows:

\begin{itemize}
    \item \textbf{Brightness}: Image brightness was globally adjusted by applying a scalar multiplier ranging from 0.05 (severe underexposure) to 3.5 (severe overexposure).
    \item \textbf{Blurring}: Three types of filters — average, Gaussian, and median — were applied with kernel sizes ranging from 3 to 11 in increments of two, introducing varying degrees of feature smoothing.
    \item \textbf{Image Resizing}: The original $32 \times 32$ images were downscaled to resolutions as low as $4 \times 4$ to assess the impact of information loss due to reduced spatial dimensions.
    \item \textbf{Other Perturbations}: Additional tests evaluated the effects of grayscale conversion, low-information images (created by embedding a $12 \times 12$ thumbnail onto a $32 \times 32$ black canvas), and the inclusion of exact and near-duplicate images. Duplicate images were introduced by retaining half of the original dataset and randomly replicating them to fill the remaining half. Near-duplicate images were generated similarly, with random adjustments to brightness, saturation, and contrast applied to the replicated subset.
\end{itemize}

A second phase of experiments examined the model's resilience to partial dataset contamination. In these tests, only a specific proportion of the training data (ranging from 4\% to 100\%) was degraded, allowing us to quantify the relationship between the prevalence of low-quality images and overall model performance.

The comprehensive results, presented in \Cref{tab:brightness} through \Cref{tab:odd-size}, reveal a clear correlation between data degradation and declining model accuracy. However, the magnitude of this effect varies substantially across different perturbation types.

An analysis of \Cref{tab:brightness} and \Cref{tab:resize} indicates that the model exhibits a degree of robustness to moderate changes in brightness and resolution. For instance, accuracy remains above 82\% for brightness scalars between 0.5 and 2.0, and for image sizes down to $17 \times 17$. This suggests that the model can generalize under moderate illumination shifts or slight reductions in spatial detail. Similarly, the presence of duplicate images or grayscale conversion resulted in only minor performance drops (\Cref{tab:other-factors}).

Conversely, blurring exerted the most pronounced negative effect on model performance (\Cref{tab:blurry}). Applying an average blur with a kernel size of just 5 caused accuracy to plummet to 54.70\%. This is likely because blurring erodes high-frequency information — such as edges and textures — that convolutional filters rely on to learn discriminative patterns. The catastrophic failure on low-information images (30.96\% accuracy) further confirms the model's dependence on clear, well-defined features.

The proportional contamination experiments (\Cref{tab:brightness-proportion} to \Cref{tab:odd-size}) reinforce these findings and highlight a critical resilience threshold. Across most degradation types, the CNN maintained accuracy above 80\% when the proportion of contaminated images was below 25\%. Beyond this threshold, performance typically declined rapidly. This suggests that while the model can tolerate a minority of poor-quality samples, its performance deteriorates sharply as these issues become more prevalent in the training data.

In conclusion, these findings quantitatively demonstrate that not all data quality issues exert equal influence. While the CNN is resilient to certain variations, it is highly sensitive to degradations that obscure or eliminate key feature information, such as blurring and severe downscaling. These results underscore the critical importance of data preprocessing and cleaning, as targeted improvements in data quality can yield substantial gains in model reliability and accuracy.

\subsection{Comparison Between the Proposed Workflow and Existing Methods on CNN Model Accuracy}
\label{sec:thresholding}

\begin{table*}[!t]
\caption{F1 Score Comparison Between Adaptive and Hard-Coded Thresholding in CleanVision Under Single Perturbations
\label{tab:cleanvision-f1}} 
\centering
\begin{threeparttable}
\begin{tabular}{|c||c|c|c|c|c|c|}
\hline
Algorithm & Light & Dark & Blurry & Low Info & Odd Size & Average \\
\hline
Original & 0.9237 & 1.0000 & 0.4735 & 0.0000 & 1.0000 & 0.6794 \\
\hline
Otsu & 0.8806 & 0.9992 & 0.6739 & 0.9889 & 1.0000 & 0.9085 \\
\hline
MET & 0.9528 & 0.9999 & 0.8243 & 0.9882 & --- & 0.9413 \\
\hline
Li & 0.9784 & 1.0000 & 0.7576 & 0.9981 & 1.0000 & 0.9468 \\
\hline
Max Entropy & 0.5170 & 0.4836 & 0.7980 & 0.8423 & 1.0000 & 0.7282 \\
\hline
GHT & 0.9185 & 0.9251 & 0.8189 & 0.9228 & 1.0000 & 0.9170 \\ 
\hline
MVE & 0.7669 & 1.0000 & 0.7171 & 0.6103 & 1.0000 & 0.8189 \\
\hline
GMM & 0.9115 & 1.0000 & 0.5826 & 0.8461 & 0.0000 & 0.6680 \\
\hline
\end{tabular}
\begin{tablenotes}[flushleft]
\footnotesize
\item \textit{Notes}: In the Odd Size adjustment, all images were resized to the same resolution, 
which resulted in only two possible scores. MET could not be executed when one of the two 
distributions had zero variance. The same limitation applies to \Cref{tab:cleanvision-f1-double}.
\end{tablenotes}
\end{threeparttable}
\end{table*}

\begin{table*}[!t]
\caption{F1 Score Comparison Between Adaptive and Hard-Coded Thresholding in CleanVision Under Two Perturbations\label{tab:cleanvision-f1-double}}
\centering
\scriptsize
\begin{adjustbox}{max width=\textwidth}
\begin{tabular}{|c||c|c|c|c|c|c|c|c|c|c|c|}
\hline
Algorithm & Light+Blurry & Light+Dark & Light+Low Info & Dark+Blurry & Dark+Low Info & Blurry+Low Info & Odd Size+Light & Odd Size+Dark & Odd Size+Blurry & Odd Size+Low Info & Average \\
\hline
Original & 0.7276 & 0.9615 & 0.6029 & 0.7937 & 0.6669 & 0.2728 & 0.9637 & 1.0000 & 0.7914 & 0.6667 & 0.7447 \\
\hline
Otsu & 0.3892 & 0.3613 & 0.3577 & 0.8186 & 0.9872 & 0.4348 & 0.5738 & 0.9994 & 0.5660 & 0.9369 & 0.6425 \\
\hline
MET & 0.2143 & 0.2143 & 0.2143 & 0.8732 & 0.9879 & 0.2143 & --- & --- & --- & --- & 0.4530 \\
\hline
Li & 0.5704 & 0.9869 & 0.9802 & 0.8737 & 0.9958 & 0.5268 & 0.9877 & 1.0000 & 0.6370 & 0.9985 & 0.8557 \\
\hline
Max Entropy & 0.2145 & 0.4255 & 0.5833 & 0.5646 & 0.5857 & 0.2163 & 0.6135 & 0.5834 & 0.8652 & 0.9856 & 0.5638 \\
\hline
GHT & 0.7270 & 0.7361 & 0.8199 & 0.7831 & 0.8649 & 0.7521 & 0.8513 & 0.8660 & 0.8259 & 0.9742 & 0.8200 \\ 
\hline
MVE & 0.5142 & 0.2013 & 0.2143 & 0.8507 & 0.6579 & 0.2209 & 0.8974 & 1.0000 & 0.3867 & 0.9506 & 0.5894 \\
\hline
GMM & 0.5476 & 0.9556 & 0.7366 & 0.7040 & 0.8373 & 0.2143 & 0.5974 & 0.6666 & 0.3308 & 0.7101 & 0.6300 \\
\hline
\end{tabular}
\end{adjustbox}
\end{table*}

\begin{table}[!t]
\caption{Comparison of F1 Scores for Near-Duplicate Detection Across Detection Methods\label{tab:nd-f1}}
\centering
\begin{threeparttable}
\begin{tabular}{|c|c|c|}
\hline
CleanVision & Fastdup & Proposed method \\
\hline
0.4579 & 0.6466 & 0.7928 \\
\hline
\end{tabular}
\end{threeparttable}
\end{table}

To validate the efficacy of our proposed workflow, we conducted a comparative analysis against existing data quality tools, CleanVision and Fastdup. The parameter settings for perturbations that yielded the lowest model accuracy in previous experiments were used to construct the evaluation datasets. Before introducing these artificial defects, we first applied CleanVision to the original, unmodified dataset. This initial scan identified and removed several blurry, overly bright, and exact duplicate images, revealing that even synthetic datasets are not as pristine as commonly assumed. The evaluation was twofold: first, we assessed various automatic thresholding algorithms for identifying general quality defects; and second, we compared performance on near-duplicate image detection.

\subsubsection{Performance of Automatic Thresholding for Quality Defect Detection}

The thresholds estimated by the aforementioned algorithms were compared against the original fixed threshold. Among these algorithms, Otsu and Li were implemented using the \texttt{scikit-image} library~\cite{van2014scikit} with default parameters, while the others were re-implemented based on the methodologies described in their respective publications. Since the number of histogram bins in Otsu's method defaults to 256, all other methods requiring bin specifications were also set to 256. The remaining parameters were configured according to the optimal settings reported in the literature.

As the primary objective was to identify low-quality images, these were treated as the positive class when computing performance metrics. Because the relative trade-off between recall and precision was not the main concern, we used the F1 score to assess overall detection effectiveness.

Our initial experiment evaluated the ability of several established thresholding algorithms to identify low-quality images in datasets containing a single type of perturbation (12\% of the images). Performance, measured by the F1 score, was benchmarked against CleanVision's original fixed-threshold method. As shown in \Cref{tab:cleanvision-f1}, most automatic thresholding methods significantly outperformed the baseline. Li's method, in particular, achieved the highest average F1 score of 0.9468, representing a 26.7\% improvement over the original method's score of 0.6794. This result underscores the inherent limitation of using a fixed threshold, which performed poorly on blurry images (0.4735) and failed entirely to detect low-information images (0.0000). While Li's method demonstrated the best overall performance, GHT also proved to be a strong alternative, exhibiting stable results across multiple perturbation types.

We then repeated the same procedure on a more complex dataset, in which two different perturbations were each applied to 6\% of the images. The advantage of dynamic thresholding became even more evident in this setting. According to the results in \Cref{tab:cleanvision-f1-double}, Li's method continued to perform best, with an average F1 score of 0.8557, again outperforming the baseline (0.7447). This scenario also revealed the weaknesses of certain methods: for example, MET, which had performed well under single perturbations, struggled with the more complex, multimodal score distributions and exhibited a sharp decline in performance. In contrast, the consistently strong results from Li and GHT highlight their robustness and suggest that they are well-suited for automating the detection of diverse and overlapping quality defects.

\subsubsection{Performance of Near-Duplicate Detection}

Beyond general quality metrics, we evaluated performance on the challenging task of near-duplicate detection. The evaluation of Fastdup involved using its built-in filtering mechanisms: for dark, bright, and blurry images, we used the default thresholds defined in the package, and for duplicates, one representative was retained per connected component. Outliers were not removed, as their detection relies on sampling a fixed proportion of images. For a fair comparison, a dedicated test set was created in which 12\% of the training images were replaced with augmented versions generated by applying random adjustments to brightness, contrast, and saturation within the range $[0.8,\ 1.2]$. This setup simulated a realistic scenario of visually similar but non-identical images.

The performance of our proposed method was compared against CleanVision and Fastdup on this task. As detailed in \Cref{tab:nd-f1}, the existing tools exhibited limited efficacy, with CleanVision and Fastdup achieving F1 scores of 0.4579 and 0.6466, respectively. In contrast, our proposed method achieved an F1 score of 0.7928, demonstrating a substantial improvement in identifying these challenging near-duplicates. This result suggests that our approach leverages a more effective feature representation and similarity measure, enabling finer distinctions between unique and redundant image content.

In summary, the empirical results confirm that our proposed workflow, which incorporates adaptive thresholding and enhanced near-duplicate detection, offers a more accurate, robust, and comprehensive solution for data cleaning compared with existing methods.



\section{Discussion and Conclusions}

This study verifies the adverse impact of data quality on machine learning within the image domain. During our experiments, we found that the FAKE subset of CIFAKE — despite being synthetic — contains not only exact duplicate pairs but also a considerable number of grayscale images, indicating that the dataset is not entirely clean. After carefully examining the algorithms implemented in existing tools, CleanVision and Fastdup, we proposed several revised detection routines and introduced dynamic threshold selection, which substantially improved the effectiveness of the light, grayscale, and near-duplicate detectors.

In addition, through a systematic evaluation of multiple thresholding algorithms, we observed that Li's Minimum Cross-Entropy Thresholding achieved the highest average performance among all methods, while Generalized Histogram Thresholding (GHT) exhibited superior stability across perturbation types. In contrast, Maximum Entropy and the Gamma Mixture Model (GMM) were less robust, and the Minimum Error Thresholding (MET) algorithm degraded markedly under dual perturbations due to the emergence of multimodal score distributions.

The results of this study are promising and pave the way toward automatically and accurately flagging problematic images in large-scale datasets. However, several avenues remain open for future investigation. First, our experiments were conducted solely on a synthetic dataset. We considered at most two concurrent perturbations and did not perform an in-depth analysis of hyperparameter selection for the thresholding algorithms. Furthermore, Li's method is sensitive to initialization (e.g., the default in \texttt{scikit-image} initializes at the global mean), which may affect its stability in other settings.

Future work will extend the proposed framework to real-world datasets, explore a broader range of quality degradations, and include comprehensive ablation studies on thresholding hyperparameters and initialization strategies. We also plan to examine additional evaluation metrics and alternative feature representations to further enhance the robustness and generalizability of automated data quality assessment tools.

\appendices
\counterwithin{table}{section}
\renewcommand{\thetable}{\thesection-\Roman{table}}
\section{Baseline CNN Configuration} \label{sec:model_used}

\begin{table}[!htbp]
\centering
\caption{The network architecture of our proposed CNN model. The model utilizes the ELU activation function and He normal weight initialization for all convolutional and dense layers, except for the final output layer which uses Softmax. Two dropout layers with a rate of 0.5 are applied in the fully-connected part of the network to mitigate overfitting}
\label{tbl:model_arch}
\begin{tabular}{@{}lll@{}}
\toprule
\textbf{Layer (Type)} & \textbf{Configuration Details} & \textbf{Output Shape} \\
\midrule
Input                 & -                              & (32, 32, 3)     \\
Conv2D\_1             & 32 filters, 7x7 kernel, ELU    & (32, 32, 32)    \\
MaxPool2D\_1           & 2x2 pool size, 2x2 stride      & (16, 16, 32)    \\
Conv2D\_2             & 64 filters, 3x3 kernel, ELU    & (16, 16, 64)    \\
Conv2D\_3             & 64 filters, 3x3 kernel, ELU    & (16, 16, 64)    \\
MaxPool2D\_2           & 2x2 pool size, 2x2 stride      & (8, 8, 64)      \\
Flatten               & -                              & (4096,)         \\
Dense\_1               & 64 units, ELU                  & (64,)           \\
Dropout\_1             & Rate = 0.5                     & (64,)           \\
Dense\_2               & 32 units, ELU                  & (32,)           \\
Dropout\_2             & Rate = 0.5                     & (32,)           \\
Dense\_3 (Output)     & 10 units, Softmax              & (10,)           \\
\bottomrule
\end{tabular}
\end{table}

\section{Empirical Analysis of Data Degradations on CNN Accuracy} \label{sec:impact_on_ml_model}

\begin{table}[!hbt]
\caption{Effect of Brightness Perturbation on CNN Classification Performance\label{tab:brightness}}
\centering
\begin{threeparttable}
\begin{tabular}{|c|c|}
\hline
\textbf{Parameter} & \textbf{Accuracy} \\
\hline
0.05 & 0.6459 \\
0.25 & 0.7952 \\
0.50 & 0.8398 \\
0.75 & 0.8744 \\
1.00 & 0.8852 \\
1.25 & 0.8721 \\
1.50 & 0.8253 \\
1.75 & 0.8427 \\
2.00 & 0.8278 \\
2.25 & 0.7411 \\
2.50 & 0.6975 \\
2.75 & 0.6909 \\
3.00 & 0.6890 \\
3.25 & 0.5999 \\
3.50 & 0.5417 \\
\hline
\end{tabular}
\begin{tablenotes}[flushleft]
\footnotesize
\item \textit{Notes}: The brightness parameter refers to the scalar applied uniformly to all pixel values in the image. For example, a value of 1.2 corresponds to multiplying every pixel in the image by 1.2, thereby increasing overall brightness.
\end{tablenotes}
\end{threeparttable}
\end{table}

\begin{table}[!hbt]
\caption{Effect of Varying Blur Kernel Sizes on CNN Classification Performance\label{tab:blurry}}
\centering
\begin{threeparttable}
\begin{tabular}{|c|c|c|c|}
\hline
\textbf{Kernel Size} & \textbf{Average Kernel} & \textbf{Gaussian Kernel} & \textbf{Median Kernel} \\
\hline
3  & 0.8073 & 0.8400 & 0.8499 \\
5  & 0.5470 & 0.7767 & 0.7791 \\
7  & 0.4603 & 0.6367 & 0.6540 \\
9  & 0.2947 & 0.5471 & 0.5101 \\
11 & 0.2581 & 0.4651 & 0.3711 \\
\hline
\end{tabular}
\end{threeparttable}
\end{table}

\begin{table}[!hbt]
\centering
\caption{Effect of Spatial Resolution Reduction on CNN Classification Accuracy}
\label{tab:resize}
\begin{tabular}{|c|c|}
\hline
\textbf{Resized size} & \textbf{Accuracy} \\
\hline
32$\times$32 (original) & 0.8852 \\
27$\times$27 & 0.8663 \\
22$\times$22 & 0.8468 \\
17$\times$17 & 0.8344 \\
12$\times$12 & 0.7936 \\
7$\times$7   & 0.7031 \\
6$\times$6   & 0.6235 \\
5$\times$5   & 0.3784 \\
4$\times$4   & 0.3210 \\
\hline
\end{tabular}
\end{table}

\begin{table}[!hbt]
\caption{Effect of Additional Data Perturbations on CNN Performance\label{tab:other-factors}}
\centering
\begin{threeparttable}
\begin{tabular}{|c|c|}
\hline
\textbf{Factor} & \textbf{Accuracy} \\
\hline
Grayscale & 0.8393 \\
Low information & 0.3096 \\
Exact duplicate & 0.8454 \\
Near duplicate (0.8--1.2) & 0.8371 \\
Near duplicate (0.5--1.5)& 0.8283 \\
\hline
\end{tabular}
\begin{tablenotes}[flushleft]
\footnotesize
\item \textit{Notes}: Near-duplicate images were generated by randomly duplicating samples and applying adjustments to brightness, contrast, saturation, and hue. Adjustment factors were uniformly sampled from either the range 0.8--1.2 or 0.5--1.5.
\end{tablenotes}
\end{threeparttable}
\end{table}

\begin{table}[!hbt]
\caption{Effect of Light/Dark Image Proportions and Brightness Scaling on CNN Accuracy\label{tab:brightness-proportion}}
\centering
\begin{threeparttable}
\begin{tabular}{|c|c|c|c|c|}
\hline
\textbf{Proportion} & \textbf{BP$= 3$} & \textbf{BP$= 2$} & \textbf{BP$=0.25$} & \textbf{BP$=0.05$} \\
\hline
0.04 & 0.8882 & 0.8855 & 0.8848 & 0.8811 \\
0.08 & 0.8759 & 0.8867 & 0.8815 & 0.8761 \\
0.12 & 0.8703 & 0.8703 & 0.8721 & 0.8610 \\
0.16 & 0.8499 & 0.8654 & 0.8702 & 0.8586 \\
0.20 & 0.8516 & 0.8499 & 0.8608 & 0.8560 \\
0.24 & 0.8305 & 0.8433 & 0.8537 & 0.8504 \\
0.25 & 0.8299 & 0.8356 & 0.8455 & 0.8404 \\
0.50 & 0.7838 & 0.8034 & 0.8217 & 0.8067 \\
0.75 & 0.6585 & 0.7101 & 0.7627 & 0.6488 \\
1.00 & 0.6890 & 0.8278 & 0.7952 & 0.6459 \\
\hline
\end{tabular}
\begin{tablenotes}[flushleft]
\footnotesize
\item \textit{Notes}: BP denotes the brightness parameter applied uniformly to the entire image. For example, BP $= 2$ indicates that all pixel values were multiplied by 2.
\end{tablenotes}
\end{threeparttable}
\end{table}

\begin{table}[!hbt]
\caption{Effect of Gaussian Blur Proportion on CNN Classification Performance\label{tab:gaussian-blur-proportion}}
\centering
\begin{threeparttable}
\begin{tabular}{|c|c|c|}
\hline
\textbf{Proportion} & \textbf{ksize$=5$} & \textbf{ksize$=7$} \\
\hline
0.04 & 0.8881 & 0.8824 \\
0.08 & 0.8853 & 0.8707 \\
0.12 & 0.8726 & 0.8726 \\
0.16 & 0.8673 & 0.8679 \\
0.20 & 0.8625 & 0.8558 \\
0.24 & 0.8557 & 0.8480 \\
0.25 & 0.8402 & 0.8481 \\
0.50 & 0.8413 & 0.8239 \\
0.75 & 0.8076 & 0.7813 \\
1.00 & 0.7767 & 0.6367 \\
\hline
\end{tabular}
\begin{tablenotes}[flushleft]
\footnotesize
\item \textit{Notes}: $k\text{size}$ refers to the kernel size used for Gaussian blurring. Larger $k$ values result in stronger blurring effects.
\end{tablenotes}
\end{threeparttable}
\end{table}

\begin{table}[!hbt]
\caption{Effect of Low-Information Image Proportion on CNN Classification Performance\label{tab:low-info}}
\centering
\begin{threeparttable}
\begin{tabular}{|c|c|}
\hline
\textbf{Proportion} & \textbf{Accuracy} \\
\hline
0.04 & 0.8823 \\
0.08 & 0.8647 \\
0.12 & 0.8584 \\
0.16 & 0.8371 \\
0.20 & 0.8457 \\
0.24 & 0.8337 \\
0.25 & 0.8213 \\
0.50 & 0.7525 \\
0.75 & 0.5053 \\
1.00 & 0.3096 \\
\hline
\end{tabular}
\end{threeparttable}
\end{table}

\begin{table}[!hbt]
\caption{Effect of Odd-Size Image Proportions on CNN Classification Performance\label{tab:odd-size}}
\centering
\begin{threeparttable}
\begin{tabular}{|c|c|c|}
\hline
\textbf{Proportion} & \textbf{12$\times$12} & \textbf{7$\times$7} \\
\hline
0.04 & 0.8792 & 0.8878 \\
0.08 & 0.8799 & 0.8786 \\
0.12 & 0.8773 & 0.8726 \\
0.16 & 0.8654 & 0.8690 \\
0.20 & 0.8659 & 0.8574 \\
0.24 & 0.8600 & 0.8530 \\
0.25 & 0.8591 & 0.8443 \\
0.50 & 0.8353 & 0.8181 \\
0.75 & 0.7991 & 0.7642 \\
1.00 & 0.7936 & 0.7031 \\
\hline
\end{tabular}
\begin{tablenotes}[flushleft]
\footnotesize
\item \textit{Notes}: Odd-sized images were generated by first downscaling to various resolutions (e.g., $12 \times 12$ and $7 \times 7$), then resizing back to $32 \times 32$, resulting in visibly distorted image content.
\end{tablenotes}
\end{threeparttable}
\end{table}

\clearpage



 
\bibliographystyle{IEEEbib}
\bibliography{IEEEabrv,references.bib}

\newpage

\end{document}